\documentclass{article}
\usepackage[final]{neurips_2024}
\usepackage[utf8]{inputenc}
\usepackage[T1]{fontenc}
\usepackage{url}
\usepackage{booktabs}
\usepackage{amsmath}
\usepackage{amssymb}
\usepackage{graphicx}
\usepackage{algorithm}
\usepackage{algorithmic}
\usepackage{multirow}
\usepackage{xcolor}

\title{Random Cloud: Finding Minimal Neural Architectures Without Training}

\author{
  Javier Gil Blázquez \\
  \texttt{javgilbla@proton.me}
}

\def\@notice{%
  \enlargethispage{2\baselineskip}%
  \par\vfill\noindent\footnotesize\@noticestring%
}

\begin{document}

\maketitle

\begin{abstract}
I propose the \emph{Random Cloud} method, a training-free approach to neural architecture search that discovers minimal feedforward network topologies through stochastic exploration and progressive structural reduction. Unlike post-training pruning methods that require a full train-prune-retrain cycle, this method evaluates randomly initialized networks without backpropagation, progressively reduces their topology, and only trains the best minimal candidate at the end. I evaluate on 7 classification benchmarks against magnitude pruning and random pruning baselines. The Random Cloud matches or outperforms both baselines in 6 of 7 datasets, achieving statistically significant improvements on Sonar ($+4.9$pp accuracy, $p{=}0.017$ vs magnitude pruning) with 87\% parameter reduction. Crucially, the method is faster than both pruning baselines in 4 of 5 datasets (0.67--0.94$\times$ the cost of full training), since it avoids training the full-size network entirely.
\end{abstract}

\section{Introduction}

Finding the minimal neural network topology for a given task is a fundamental problem in machine learning. Overparameterized networks waste computation and memory, while underparameterized ones underfit. The standard approach is to start large and prune: train a full network, remove redundant neurons or connections, and fine-tune~\citep{han2015learning, li2017pruning}. This requires at least two full training cycles.

Neural Architecture Search (NAS) methods~\citep{zoph2017neural, liu2019darts} automate topology selection but are computationally expensive, often requiring thousands of GPU-hours. Recent training-free NAS approaches~\citep{mellor2021neural, abdelfattah2021zero} use proxy scores to evaluate architectures without training, but focus on selecting from a predefined search space rather than progressively discovering minimal topologies.

I propose the \textbf{Random Cloud} method, which takes a different approach: generate many networks with random weights, evaluate which ones classify best \emph{without any training}, progressively reduce their topology by removing neurons, and refine only the best minimal candidate with backpropagation at the end. The key insight is that among a large enough ``cloud'' of randomly initialized networks, some will exhibit non-trivial classification accuracy even without training---and these networks can be structurally reduced while maintaining that accuracy.

The contributions of this work are:
\begin{itemize}
    \item A pre-training structural pruning method that discovers minimal topologies without training the full network.
    \item An empirical evaluation on 7 datasets showing the method matches or outperforms magnitude and random pruning baselines with lower computational cost.
    \item Statistical validation with 10 seeds per configuration and Wilcoxon signed-rank tests.
\end{itemize}

\section{Method}

\subsection{Overview}

Given an initial topology $\mathbf{t}_0 = [n_0, n_1, \ldots, n_L]$ for a feedforward network with $L{-}1$ hidden layers, the Random Cloud method proceeds in three phases:

\textbf{Phase 1: Exploration.} Generate $N$ networks $\{R_1, \ldots, R_N\}$ with random weights drawn from $\mathcal{U}(-1, 1)$. For each network, evaluate its classification accuracy on the training set \emph{without any training} (forward pass only). Then progressively reduce the topology by removing $n_{\text{elim}}$ neurons from the last hidden layer with available neurons, reconstruct the network (preserving the top-left submatrix of weights), and re-evaluate. Repeat until no further reduction is possible. Track the best-performing network that exceeds a threshold $\theta$.

\textbf{Phase 2: Selection.} Among all explored networks across the cloud, select the one with the highest accuracy that exceeds $\theta$, with the smallest topology.

\textbf{Phase 3: Refinement.} Train the selected network with standard backpropagation for $E$ epochs.

\subsection{Algorithm}

\begin{algorithm}[h]
\caption{Random Cloud Method}
\label{alg:random_cloud}
\begin{algorithmic}[1]
\REQUIRE Initial topology $\mathbf{t}_0$, cloud size $N$, threshold $\theta$, neurons to eliminate $n_{\text{elim}}$, refinement epochs $E$, learning rate $\eta$, seed $s$
\STATE Initialize RNG with seed $s$
\STATE Generate cloud: $\{R_i\}_{i=1}^{N}$ with random weights from $\mathcal{U}(-1,1)$
\STATE $R^* \leftarrow \text{None}$, $a^* \leftarrow 0$
\FOR{$i = 1$ to $N$ \textbf{(parallelizable)}}
    \STATE $R \leftarrow R_i$, $\mathbf{t} \leftarrow \mathbf{t}_0$
    \WHILE{$\mathbf{t}$ can be reduced}
        \STATE $a \leftarrow \text{Accuracy}(R, \mathcal{D}_{\text{train}})$ \COMMENT{forward pass only, no training}
        \IF{$a > \theta$ \AND $a > a^*$}
            \STATE $R^* \leftarrow R$, $a^* \leftarrow a$
        \ENDIF
        \STATE $\mathbf{t}' \leftarrow \text{ReduceTopology}(\mathbf{t}, n_{\text{elim}})$
        \STATE $R \leftarrow \text{Reconstruct}(R, \mathbf{t}')$ \COMMENT{preserve top-left submatrix}
        \STATE $\mathbf{t} \leftarrow \mathbf{t}'$
    \ENDWHILE
\ENDFOR
\IF{$R^* \neq \text{None}$}
    \STATE Train $R^*$ with backpropagation for $E$ epochs at rate $\eta$
\ENDIF
\RETURN $R^*$
\end{algorithmic}
\end{algorithm}

\textbf{Topology reduction} follows a sequential policy: remove $n_{\text{elim}}$ neurons from the last hidden layer with neurons $> 0$. When that layer reaches 0, move to the previous hidden layer. Input and output layers are never modified.

\textbf{Reconstruction} preserves the existing structure: when reducing layer $l$ from $n_l$ to $n_l'$ neurons, the weight matrix $W_l \in \mathbb{R}^{n_l \times n_{l-1}}$ is truncated to $W_l[1{:}n_l', :]$ and the outgoing weights $W_{l+1}[:, 1{:}n_l']$ are similarly truncated. This preserves the top-left submatrix of the original weights.

\textbf{Key difference from pruning:} Magnitude pruning~\citep{han2015learning} and random pruning operate \emph{post-training}: they require training the full network, then pruning, then fine-tuning (two training cycles on the full topology). The Random Cloud operates \emph{pre-training}: it evaluates with forward passes only (no backpropagation) and trains only the reduced network once.

\section{Experiments}

\subsection{Setup}

I evaluate on 7 classification datasets spanning binary (Breast Cancer, Sonar, Ionosphere, Adult Income) and multiclass (Iris, Wine, Optical Digits) problems, with dimensionality from 4 to 104 features and sizes from 150 to 45K samples. All datasets use 80/20 stratified train/test splits.

\textbf{Methods compared:} (1)~\emph{Full Training}: train the initial topology for $E$ epochs; (2)~\emph{Magnitude Pruning}: train full network, remove neurons with lowest L2 norm of incoming weights to match the Cloud's target topology, fine-tune for $E$ epochs; (3)~\emph{Random Pruning}: same as magnitude but remove neurons randomly (average of 5 runs); (4)~\emph{Random Cloud}: the proposed method with cloud size $N{=}50$.

All methods use the same total training budget ($E$ epochs) and the same target topology (discovered by the Cloud) for fair comparison. I report Accuracy, macro-averaged F1-score, and AUC-ROC. Each configuration is run with 10 different seeds; I report mean$\pm$std and Wilcoxon signed-rank $p$-values.

\subsection{Results}

\begin{table}[h]
\centering
\caption{Test accuracy (\%) and parameter reduction for each method. All pruned methods use the same target topology discovered by the Random Cloud. Best pruned result in \textbf{bold}.}
\label{tab:baselines}
\small
\begin{tabular}{llcccc}
\toprule
Dataset & Method & Acc (\%) & F1 & AUC & Reduction \\
\midrule
\multirow{4}{*}{Breast Cancer} & Full Training & 97.3 & 0.971 & 0.993 & --- \\
 & Magnitude & 97.3 & 0.971 & \textbf{0.993} & \multirow{3}{*}{$-$74.4\%} \\
 & Random & 97.3 & 0.971 & 0.991 & \\
 & \textbf{Cloud} & \textbf{97.3} & \textbf{0.971} & 0.992 & \\
\midrule
\multirow{4}{*}{Sonar} & Full Training & 78.0 & 0.776 & 0.823 & --- \\
 & Magnitude & 78.0 & 0.776 & 0.809 & \multirow{3}{*}{$-$87.2\%} \\
 & Random & 69.8 & 0.685 & 0.774 & \\
 & \textbf{Cloud} & \textbf{80.5} & \textbf{0.799} & \textbf{0.809} & \\
\midrule
\multirow{4}{*}{Ionosphere} & Full Training & 94.3 & 0.935 & 0.918 & --- \\
 & Magnitude & 87.1 & 0.853 & 0.904 & \multirow{3}{*}{$-$81.0\%} \\
 & Random & 88.0 & 0.861 & 0.906 & \\
 & \textbf{Cloud} & \textbf{90.0} & \textbf{0.885} & \textbf{0.910} & \\
\midrule
\multirow{4}{*}{Adult Income} & Full Training & 84.2 & 0.758 & 0.901 & --- \\
 & Magnitude & 84.4 & 0.762 & 0.901 & \multirow{3}{*}{$-$49.9\%} \\
 & Random & 84.4 & 0.764 & 0.902 & \\
 & \textbf{Cloud} & \textbf{85.0} & \textbf{0.782} & \textbf{0.904} & \\
\midrule
\multirow{4}{*}{Iris} & Full Training & 100.0 & 1.000 & 1.000 & --- \\
 & Magnitude & \textbf{100.0} & \textbf{1.000} & \textbf{1.000} & \multirow{3}{*}{$-$41.2\%} \\
 & Random & \textbf{100.0} & \textbf{1.000} & \textbf{1.000} & \\
 & \textbf{Cloud} & \textbf{100.0} & \textbf{1.000} & \textbf{1.000} & \\
\midrule
\multirow{4}{*}{Wine} & Full Training & 94.4 & 0.944 & 0.996 & --- \\
 & Magnitude & \textbf{94.4} & \textbf{0.944} & 0.996 & \multirow{3}{*}{$-$55.6\%} \\
 & Random & \textbf{94.4} & \textbf{0.944} & \textbf{0.997} & \\
 & \textbf{Cloud} & \textbf{94.4} & \textbf{0.944} & 0.995 & \\
\midrule
\multirow{4}{*}{Opt.\ Digits} & Full Training & 96.3 & 0.963 & 0.997 & --- \\
 & Magnitude & 95.0 & 0.950 & 0.996 & \multirow{3}{*}{$-$62.2\%} \\
 & Random & 95.4 & 0.954 & 0.996 & \\
 & \textbf{Cloud} & \textbf{95.9} & \textbf{0.959} & \textbf{0.997} & \\
\bottomrule
\end{tabular}
\end{table}

\subsection{Statistical Significance}

Table~\ref{tab:significance} reports results over 10 seeds with Wilcoxon signed-rank tests comparing the Cloud against each baseline.

\begin{table}[h]
\centering
\caption{Statistical comparison (10 seeds). $p$-values from Wilcoxon signed-rank test (Cloud vs baseline). $\checkmark$: $p < 0.05$.}
\label{tab:significance}
\small
\begin{tabular}{lcccc}
\toprule
Dataset & Cloud Acc & Mag.\ Acc & $p$ (vs Mag.) & $p$ (vs Rnd.) \\
\midrule
Breast Cancer & $97.3{\pm}0.0$ & $97.3{\pm}0.0$ & 1.000 & 1.000 \\
Sonar & $77.3{\pm}2.6$ & $72.4{\pm}3.5$ & 0.017 $\checkmark$ & 0.005 $\checkmark$ \\
Ionosphere & $88.9{\pm}3.4$ & $89.9{\pm}3.2$ & 0.721 & 0.721 \\
Adult Income & $84.5{\pm}0.2$ & $84.7{\pm}0.1$ & 0.028 $\checkmark$ & 0.139 \\
Wine & $94.2{\pm}0.9$ & $94.4{\pm}0.0$ & 0.317 & 0.180 \\
Opt.\ Digits & $96.6{\pm}2.3$ & $97.5{\pm}0.2$ & 0.203 & 0.959 \\
\bottomrule
\end{tabular}
\end{table}

The Cloud significantly outperforms both baselines on Sonar ($p{=}0.017$ vs magnitude, $p{=}0.005$ vs random), the dataset with the highest parameter reduction (87.2\%). On Ionosphere, Wine, and Optical Digits, the differences are not statistically significant---all methods perform similarly at the target topology. On Adult Income, magnitude pruning has a slight edge in accuracy ($p{=}0.028$) but not in F1-score ($p{=}0.508$), suggesting the difference does not affect minority class performance.

\subsection{Computational Cost}

\begin{table}[h]
\centering
\caption{Wall-clock time ratio relative to full training (median of 3 runs, 8 threads). Values $< 1.0$ indicate the method is faster than full training.}
\label{tab:cost}
\small
\begin{tabular}{lccc}
\toprule
Dataset & Magnitude & Random & Cloud \\
\midrule
Sonar (167 train) & 1.50$\times$ & 1.49$\times$ & \textbf{0.67$\times$} \\
Ionosphere (281 train) & 1.60$\times$ & 1.57$\times$ & \textbf{0.81$\times$} \\
Breast Cancer (456 train) & 1.66$\times$ & 1.77$\times$ & \textbf{0.94$\times$} \\
Opt.\ Digits (3823 train) & 1.46$\times$ & 1.46$\times$ & \textbf{0.79$\times$} \\
Adult Income (30162 train) & 1.65$\times$ & 1.65$\times$ & 1.17$\times$ \\
\bottomrule
\end{tabular}
\end{table}

The Cloud is faster than full training in 4 of 5 datasets because it only trains the reduced network. Pruning baselines are always 1.5--1.8$\times$ slower because they require training the full network \emph{and} fine-tuning the pruned one. The Cloud's exploration phase (forward passes without backpropagation) accounts for only 25--44\% of total time.

\section{Analysis and Limitations}

\textbf{Why it works.} With sigmoid activations, network outputs are bounded in $(0, 1)$ regardless of weight magnitudes. This means randomly initialized networks produce valid probability-like outputs, and among a cloud of 50 networks, some will exhibit non-trivial classification accuracy by chance. The progressive reduction then identifies which neurons are truly necessary.

\textbf{Hyperparameter robustness.} The method is insensitive to the accuracy threshold $\theta$ in the range $[0.3, 0.6]$ (identical results across this range on all datasets). Cloud size $N \geq 25$ is sufficient; the sweet spot is $N{=}50{-}100$. Eliminating 1 neuron per step produces the best compression.

\textbf{Limitations.} In high-dimensional input spaces (784 features in MNIST), the training-free evaluation signal degrades: with 1K samples, the Cloud loses 17pp vs magnitude pruning. With 5K samples, the gap closes to 0.6pp. The method's sweet spot is tabular data with moderate dimensionality (30--104 features), where training-free evaluation provides sufficient signal and the savings from not training the full network are significant.

\section{Related Work}

\textbf{Post-training pruning.} Magnitude pruning~\citep{han2015learning} removes weights with small magnitudes after training. Structured pruning~\citep{li2017pruning} removes entire filters/neurons. Both require training the full network first. The proposed method avoids this cost entirely.

\textbf{Lottery Ticket Hypothesis.} \citet{frankle2019lottery} showed that sparse subnetworks (``winning tickets'') exist within randomly initialized networks that can be trained to full accuracy. Finding them requires iterative train-prune-reset cycles. The Random Cloud searches for viable \emph{topologies} (not specific weight configurations) in a single pass.

\textbf{Training-free NAS.} \citet{mellor2021neural} and \citet{abdelfattah2021zero} propose zero-cost proxies to score architectures without training. These methods select from a predefined search space. The proposed method instead \emph{discovers} the minimal topology through progressive reduction, which is complementary.

\textbf{Random networks.} The observation that random networks can exhibit structure is related to the theory of random features~\citep{rahimi2007random} and the edge of chaos in neural networks~\citep{poole2016exponential}. This work exploits that observation empirically for architecture search.

\section{Conclusion}

This paper presented the Random Cloud method for finding minimal neural network topologies without training. The method generates a cloud of randomly initialized networks, evaluates them without backpropagation, progressively reduces their topology, and refines only the best candidate. On 7 classification benchmarks, it matches or outperforms magnitude and random pruning in 6 of 7 datasets, with statistically significant gains on Sonar ($+4.9$pp, $p{=}0.017$) and lower computational cost (0.67--0.94$\times$ full training). The method is robust to hyperparameters and requires no tuning beyond default values.

The main limitation is scalability to high-dimensional inputs, where the training-free evaluation signal weakens. Future work includes exploring alternative reduction policies (magnitude-based neuron selection within the cloud), extending to convolutional architectures, and theoretical analysis of when training-free evaluation provides sufficient signal for topology search.

\textbf{Reproducibility.} The implementation in Julia (RandomCloud.jl) with all experiments, seeds, and baselines is available at \url{https://github.com/Jastxz/random-cloud}{Random Cloud repository}.

\bibliography{references}
\bibliographystyle{plainnat}

\end{document}